\documentclass[journal]{IEEEtran}
\usepackage{color}
\usepackage[colorinlistoftodos]{todonotes}
\usepackage[margins]{changes}
\definechangesauthor[color=red]{VS}
\usepackage{subcaption}

\ifCLASSINFOpdf
\else
\fi
\usepackage{graphicx}
\graphicspath{{./images/}}
\usepackage{wrapfig}
\usepackage[rightcaption]{sidecap}

\usepackage{float}

\usepackage[hyphens]{url}

\usepackage[
bookmarksopen,
bookmarksdepth=2,
colorlinks,
citecolor=teal,
citebordercolor=teal,
urlcolor=blue]{hyperref}

\usepackage[affil-it]{authblk}

\begin{document}
%
\title{Benchmark Dataset for Automatic Damaged Building Detection from Post-Hurricane Remotely Sensed Imagery}
%
%
%

%


\author[1]{Sean~Andrew~Chen$^{\dagger}$}
\author[2]{Andrew~Escay$^{\dagger}$}
\author[3]{Christopher~Haberland$^{\dagger}$}
\author[4]{Tessa~Schneider$^{\dagger}$}
\author[3]{Valentina Staneva}
\author[3]{~Youngjun~Choe$^{*}$\thanks{$^{\dagger}$The four authors (in alphabetical order by last names) equally contributed to this work. $^{*}$Corresponding author (e-mail: ychoe@u.washington.edu).}}
\affil[1]{New York University}
\affil[2]{University of the Philippines Diliman}
\affil[3]{University of Washington}
\affil[4]{German Corporation for International Cooperation GmbH (GIZ)}

\markboth{}
{Shell \MakeLowercase{\textit{et al.}}: Bare Demo of IEEEtran.cls for IEEE Journals}
%



\maketitle

\begin{abstract}

Rapid damage assessment is of crucial importance to emergency responders during hurricane events, however, the evaluation process is often slow, labor-intensive, costly, and error-prone. New advances in computer vision and remote sensing open possibilities to observe the Earth at a different scale. However, substantial pre-processing work is still required in order to apply state-of-the-art methodology for emergency response. To enable the comparison of methods for automatic detection of damaged buildings from post-hurricane remote sensing imagery taken from both airborne and satellite sensors, this paper presents the development of benchmark datasets from publicly available data. The major contributions of this work include (1) a scalable framework for creating benchmark datasets of hurricane-damaged buildings and (2) public sharing of the resulting benchmark datasets for Greater Houston area after Hurricane Harvey in 2017. The proposed approach can be used to build other hurricane-damaged building datasets on which researchers can train and test object detection models to automatically identify damaged buildings.
\end{abstract}



%
\IEEEpeerreviewmaketitle

\section{Introduction} 
%
%
%
%
\IEEEPARstart{E}{mergency} managers of today grapple 
with post-hurricane damage assessment that largely relies on field surveys and damage reports \cite{FEMA2016}. The recent expansion of private and government satellite imaging operations and their push to share some of the acquired data presents new opportunities for observing hurricane affected areas \cite{duda2011usgs}. 
New methods in processing aerial and satellite images have improved assessment efficiency, but the process still depends on human visual inspection \cite{ofli2016combining,barrington2011crowdsourcing}. In the aftermath of Hurricane Irma in 2017, analysts at the U.S. National Geospatial-Intelligence Agency sifted through hundreds of satellite images each day for damage assessment \cite{simonite_2018}. These labor-intensive approaches are expensive and inefficient \cite{mnih_2013}. Further, delayed assessment slows down urban search and rescue response times \cite{pohl_2018}.


Despite the availability of various disaster-relevant public data, they are not always in a format to easily access, integrate and process.
This paper presents an important first step towards the automatic detection of damaged buildings on post-hurricane remote sensing imagery taken from both airborne and satellite sensors.  In our work we propose a scalable framework to create benchmark datasets of hurricane-damaged buildings from terabytes of data. We also publicly share the resulting benchmark datasets for Greater Houston area after Hurricane Harvey, 2017.
    The benchmark datasets are suitable for training and testing of state-of-the-art \textit{object detection} models which have been already successful in detecting objects from various categories in other domains. 
    
    Such benchmark data development effort is called for by machine learning researchers in the remote sensing domain \cite{cheng2016survey}. For example, benchmark datasets for aerial scene classification are widely used \cite{xia2017aid}. A benchmark dataset for damaged-building \textit{classification} is also developed recently \cite{cao2018deep} and is distinct from this work because data for classification cannot be used for object detection that requires localization of an object of interest in addition to its classification to a correct category. 

    
    
    Our benchmark datasets consist of raster (satellite and aerial imagery) and vector data (auxiliary building damage information), which together provide the necessary components to train a machine learning model. The vector data, including crowdsourced damage annotations from the TOMNOD project (\url{https://www.tomnod.com/}),  
    flood damage estimates by the U.S. Federal Emergency Management Agency (FEMA), and bounding boxes, are shared publicly (see Appendix). The raw raster data (in order of terabytes), shared by DigitalGlobe and the U.S. National Oceanic and Atmospheric Administration (NOAA), are available through the stable URLs of the original data sources as described later. The data contains RGB bands only. 
    
    The remainder of this paper is organized as follows. Section~\ref{sec:background} summarizes existing work on disaster damage assessment using satellite imagery. Section~\ref{sec:data} details the process of creating the benchmark dataset. 
    Section~\ref{sec:conclusion} concludes the paper with remarks on future research directions.

\section{Background} \label{sec:background} 


Current damage assessment methods for emergency managers consist largely of field or windshield surveys and damage reports
\cite{FEMA2016, yang2011hierarchical, hristidis2010survey}.  Interviews with emergency managers reveal that this practice requires significant information integration resources.  Aerial imagery is becoming increasingly more pervasive in damage assessment practice since it can be captured and processed within hours, while satellite imagery could take days
\cite{ofli2016combining}.  A few studies directly compare aerial and satellite imagery for assessment reliability, finding satellite imagery to be useful for damage pattern recognition \cite{corbane2011comparison}, and aerial imagery to be helpful for estimation of the intensity of building damages
\cite{lemoine2013intercomparison}.  

However, the current capability of satellite data collection and availability is improving.  International organizations (e.g. United Nations Platform for Space-based Information for Disaster Management and Emergency Response, International Charter on Space and Major Disasters) and national agencies (e.g. NASA, USGS, NOAA) are sharing satellite imagery to aid damage assessment 
\cite {tralli2005satellite, duda2011usgs}.  Commercial satellite imagery companies (e.g. DigitalGlobe, Planet Labs) are releasing pre- and post-event satellite imagery
\cite{digitalglobe2018, planet2018}, and other organizations are releasing real-time satellite imagery in the US and Europe
\cite{bielski2011post, earthnow2018}.




The use of automatic damage detection systems that take satellite imagery as input is uneven across different types of natural disasters. While automatic methods for earthquake damage assessment are relatively well-established \cite{bialas2016object}, they are less so for hurricane damage assessment.  Within the domain of hurricanes, flood detection remains the focus of existing methods, which leaves other types of damages, such as wind-induced ones, neglected.  Synthetic-aperture radar (SAR) images are typically used for this task \cite{dfo2018, science2018}.  Segmentation has been used to automatically annotate flooded roads where pre- and post-event satellite imagery is available
\cite{medium2017}.  
Other flood detection methods utilize certain spectral bands, namely near-infrared in optical sensor images
\cite{gbdx2018, planet2} to detect impure water, a proxy for a flooded area.  These models rely on a selected threshold that is dependent on factors such as time of day and geographical characteristics.
The reliance on such thresholds limits the generalizability of this model to new events
\cite{limmer2016infrared}. The most prominent work for both earthquake and hurricane damage assessment is the Advanced Rapid Imaging and Analysis (ARIA) project, which uses SAR sensor outputs based on a physics-based understanding of the way damages appear on SAR images 
\cite{aria2018}.  In contrast to the ARIA project, this work focuses on creating a dataset and prepares it for statistical machine learning of damages of any type recognizable by humans on pansharpened satellite images from optical sensors.

Many existing (semi-)automated damage assessment methods using satellite imagery take either physics-based or rule-based approaches
\cite{dong2013comprehensive}.  Methods on optical images extract and use various properties of damages from images
\cite{rathje2005damage}.  These are fine-tuned to a particular event and although they appear effective for a past event, these are not applicable to other events
\cite{bignami2011objects}.  Some methods require pre-event imagery for comparison with post-event imagery.  These, too, are less generalizable to other events, especially to those in regions where pre-event imagery is not available
\cite{li2011improved}.  Methods using SAR imagery have even more limited generalizability than those using optical imagery due to the small archive of SAR imagery that is available
\cite{brunner2010earthquake}.  Another damage assessment method, classification, has been used to determine whether damaged buildings occur in satellite imagery \cite{cao2018deep}. 
Object detection allows for the identification and localization of multiple object classes, such as the 60 classes (e.g. passenger vehicle, fixed-wing aircraft, building) defined in the Xview dataset, one of the largest publicly available overhead imagery object detection datasets.  According to Lam et al. (2018), ``several object detection datasets exist in the
natural imagery space, but there are few for
overhead satellite imagery''
\cite{lam2018xview}.  Planet Labs has also recently developed a training dataset using crowdsourced annotations on satellite images, which were then chipped and visually inspected.  The dataset includes an ontology of objects found in regions affected by disasters prepared for object detection \cite{Erinjippurath2018}.  This dataset has not been shared publicly.




Building on the momentum of the public Xview dataset, this paper discusses the preparation of a public dataset using post-event satellite imagery from optical sensors, as well as, aerial imagery.  This dataset was developed for training object detection models.  

    \begin{figure*}

      \centering
        \includegraphics[width=1\textwidth]{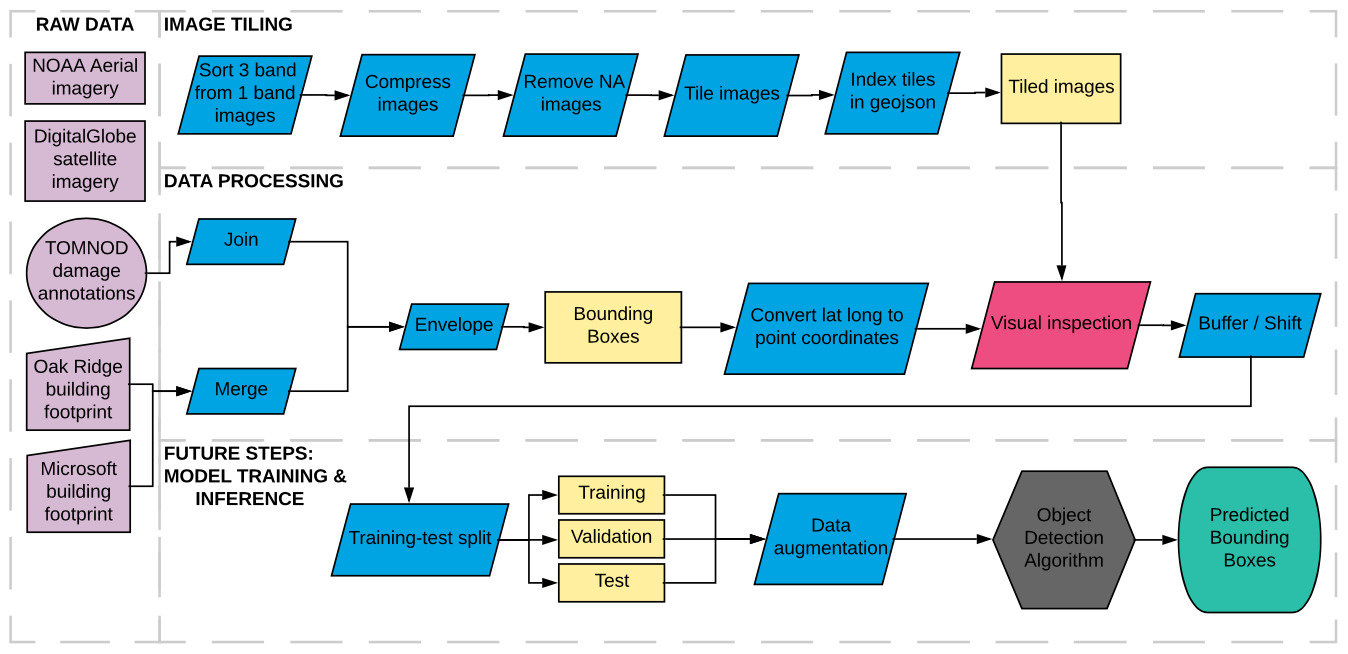}
          \caption{\textbf{Benchmark Dataset Preparation Process.} In the above diagram we describe the steps of creating a benchmark dataset: the first row indicates the preprocessing steps which are required to convert the large raw datasets to a more manageable tiled format; the second row describes how the damage annotation vector data is joined with the raster data to obtain corresponding bounding boxes; the last row illustrates a traditional workflow that a machine learning practitioner will take to train object detection algorithms on the resulting benchmark dataset.}
    \end{figure*}

\section{Benchmark Data Development}\label{sec:data} 

We clean and compile two new data sets that contain observations of both damaged and non-damaged buildings within post-Harvey images for input into an object detection model. Our approach is characterized by the following process: 

\begin{itemize}
\item Obtain data identifying buildings that visually appear to be damaged from satellite imagery.
\item Join the damaged annotations to a comprehensive building footprint dataset to record 
\item Create minimum-bounding envelopes around the building footprints to provide accurate class-distinguished bounding boxes for object detection modeling.
\end{itemize}

%

\subsection{Data Criteria \& Challenges}

	There are several criteria when sourcing data for identifying damaged buildings with an object detection model. First, we must consider its verifiability as representing buildings that have actually sustained damaged from the natural disaster in question. We also desire that the data represent damage in a visually apparent way, so as not to confuse the object detection model, which relies upon RGB bands to learn features of damaged buildings in imagery. In reality, it is possible that a building be verified by inspectors as having sustained catastrophic interior or structural damage yet provide few distinguishing visual cues via satellite imagery. Another desirable trait for source data is comprehensiveness. Failure to accurately account for all input data that should be classified according to the target classification schema injects noise into the model training. A final desirable trait is functionality: can the data be manipulated into the proper format for input into an object detection model.
    
    In addition, we have deliberately focused on the use of only RGB spectral bands. Many imagery satellites carry instruments for multiple bands beyond RGB. However, by relying on RGB, we have made speed and cost of imagery acquisition - constraints that emergency responders may very well face after a disaster - priorities. Aerial imagery from planes or drones will often be limited in their instrument payload capabilities, making reliance on RGB even more critical. Furthermore, RGB imagery is also advantageous for visual inspection: if an emergency procedure requires emergency responders to verify some of the damage predictions before actions are taken, it is much easier for the untrained eye to inspect the RGB bands as opposed to bands from the invisible spectrum, which have unnatural colors.

    For the object detection model to work well, the data should be maximally informative while it observes the considerations outlined above.

    \subsection{Data Sources}
    We identify two sources of annotation data that satisfy these criteria to different degrees: 1. crowdsourced annotation data from the TOMNOD project \cite{TOMNOD} identifying damaged buildings from DigitalGlobe satellite imagery \cite{digitalglobe2018} and 2. data collected by FEMA \cite{FEMAdata} documenting damaged property parcels following Hurricane Harvey identifying five classes of building damages. In addition to the annotation data, we identified and employed two different imagery datasets: 1. open sourced post-Harvey imagery captured and hosted by DigitalGlobe, the parent company of TOMNOD and 2.  the NOAA aerial imagery survey \cite{noaa2018} conducted immediately after the disaster event. The disparate datasets will hereafter be referred to by the names of the organizations that created them (TOMNOD, FEMA, DigitalGlobe, NOAA).
    
    A final benchmark dataset ultimately requires one set of raster imagery paired with vectors of building bounding boxes annotated according to their damage status. These four initial datasets were processed and evaluated to obtain the optimal final dataset configuration. After these data explorations, we came to two main configurations: 1. TOMNOD data paired with DigitalGlobe imagery and 2. FEMA data paired with NOAA imagery. Each dataset contained its own benefits and drawbacks. Ultimately, however, we have chosen the FEMA and NOAA dataset. 
    
    The imagery and annotation data - however - were not the only datasets used. Intermediate products were used to properly process them into useful final products. Building footprints were critically important. We relied on two main datasets: the open source Microsoft national building footprint dataset \cite{MicrosoftBF} and the Oak Ridge National Laboratory dataset \cite{OakRidgeBF} of the Harvey affected areas created specifically for Harvey disaster response. Again, these two datasets had different benefits and drawbacks. 
    
    Finally, in some cases, parcel data from the various affected counties was utilized in the creation of the final dataset as well. While some of these parcel datasets were proprietary and required permission from an individual county, many were open source.
    
    We will now describe in further detail how we processed each dataset in creating our final dataset.

    \subsection{General Processing Steps}
    In order for an object detection algorithm to properly learn from a dataset, it must be able to differentiate between different object classes. In this case, the algorithm must learn the difference between damaged and non-damaged buildings. Our first challenge is then to go from a single annotation point to the outline of an actual building. Because the algorithm may learn from the objects in direct proximity to other factors, it is actually ideal to go from the building footprint to a bounding box of the building - the least area rectangle fully surrounding the structure. 
    
    The remote sensing imagery - due to its large file size - is tiled into smaller sets of images which are more easily handled. As these images are georeferenced, they can be overlaid with the bounding box information, from which the algorithm learns.

    \subsection{TOMNOD \& DigitalGlobe} 
    
    The original TOMNOD vector data was retreived in June 2018 \cite{TOMNOD}. This dataset was created from crowdsourced volunteers who were prompted to view DigitalGlobe satellite imagery of the Hurricane Harvey aftermath captured on 9-5-2018 and demarcate geolocated points identifying an apparently damaged structure. The TOMNOD data contains four classes of damage annotations: damaged buildings, damaged roads, damaged bridges, and trash heaps. We extracted only the 18474 damaged building annotations from this full set, as the quantity of non-building categories was insufficient as input for current deep-learning-based object detection models. The TOMNOD building annotations were located in 19 counties across southeastern Texas and western Louisiana, with the majority in Houston in Harris County, Texas. 
    
    It is important to note that due to the nature of the annotation task being carried out by volunteers who are untrained, as well as the difficulty of exactly defining the location of points from high resolution imagery, points from the TOMNOD dataset are often inexactly referenced to buildings in RGB satellite imagery. Visual inspection of the data confirmed that some points did not rest on structures in the DigitalGlobe imagery, although they generally were located in the general vicinity of an identifiable structure. Additionally, several annotators may have been presented the same imagery, leading to clusters of points around the same structure reflecting multiple annotations for the same structure. Because of this, a data creation process of storing bounding boxes around each point annotated by volunteers would lead to duplicate data.
\begin{figure}
    \begin{subfigure}{\linewidth}
      \centering
        \includegraphics[width=\linewidth]{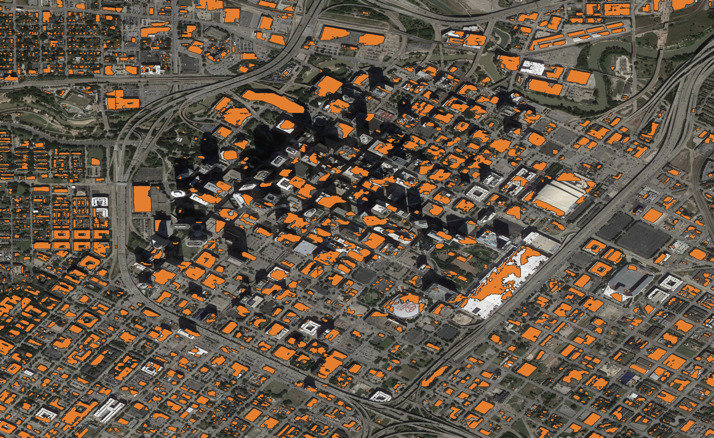}
              \caption{A sample image from Oak Ridge National Laboratory building footprints.}\label{fig:oak}
    \end{subfigure}
    
    \begin{subfigure}{\linewidth}
      \centering
        \includegraphics[width=\linewidth]{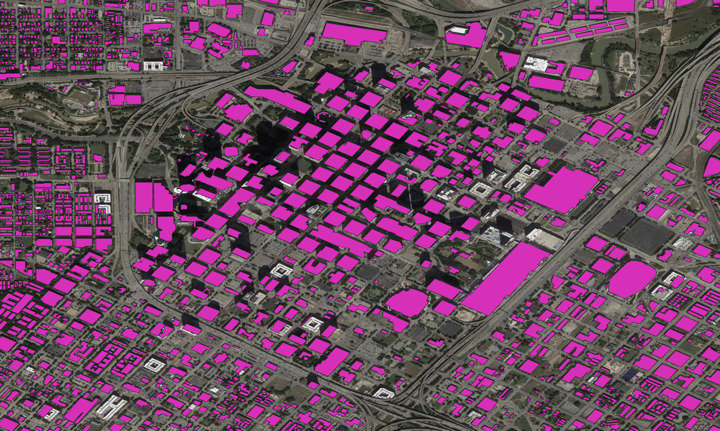}
              \caption{A sample image from Microsoft's building footprints.}\label{fig:msft}
    \end{subfigure}
    \caption{Building footprints created through different approaches: one can notice that the Microsoft's building footprints look more precise than the Oak Ridge ones}
    \end{figure}
    
    To circumvent these problems, we joined each TOMNOD point representing the incidence of a damaged building with a building footprint polygon. The building footprint data was obtained from both Microsoft \cite{MicrosoftBF} and Oak Ridge National Laboratory\cite{OakRidgeBF}. While Microsoft building footprint data did not provide full coverage of the study area, Oak Ridge National Laboratory building footprints did. However, Oak Ridge National Laboratory building footprints were often of irregular shape (see Figure~\ref{fig:oak}) and less accurate than Microsoft building footprints (see Figure~\ref{fig:msft}). Although the Oak Ridge National Laboratory building footprint dataset is spatially comprehensive for the study area, it contains many polygons that are falsely recognized as buildings. We clean it by eliminating contiguous polygons that are less than 16 square meters in area. This is a heuristic to ensure that bounding boxes were not created around false-positive data or extremely small structures. To maximize coverage and building accuracy from available resources, we create a joined building footprint dataset that consists of the union of building footprint data freely released by Microsoft and non-intersecting building footprint data from Oak Ridge National Laboratory. TOMNOD damage annotations were joined to the closest building within .00015 degrees (approximately 16 meters). A minimum bounding rectangle was then created around each building footprint polygon, constituting the final bounding box.
     Because the TOMNOD data did not contain exact reference points of the geographic bounds of satellite imagery within which buildings were visually inspected by volunteers from above, the buildings that should be considered ``non-damaged" were estimated as covering all building footprints that were not joined to TOMNOD points in tiled DigitalGlobe images that also contained TOMNOD points.

     \begin{figure}
 
      \centering
        \includegraphics[width=0.5\textwidth]{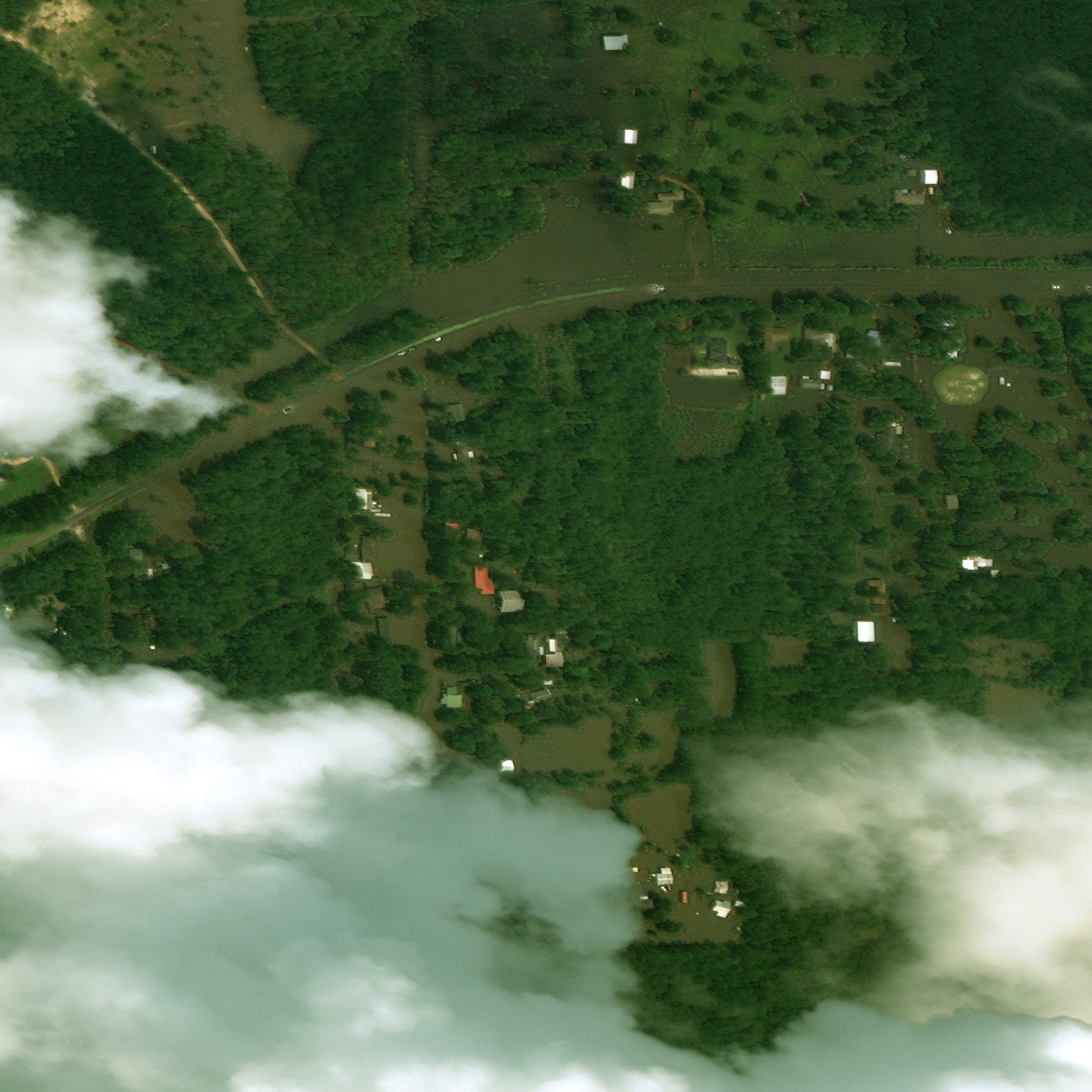}
    \caption{A sample image tile from DigitalGlobe.}
    \end{figure}

     \subsection{FEMA \& NOAA}

   \begin{figure}
      \centering
        \includegraphics[width=0.5\textwidth]{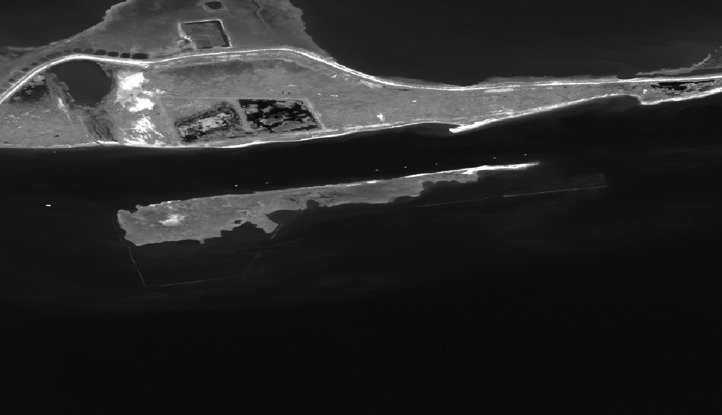}
        
          \caption{A sample image tile from DigitalGlobe where the other color bands are missing.}
          \label{fig:bw}
    \end{figure}

    \begin{figure}
      \centering
        \includegraphics[width=0.5\textwidth]{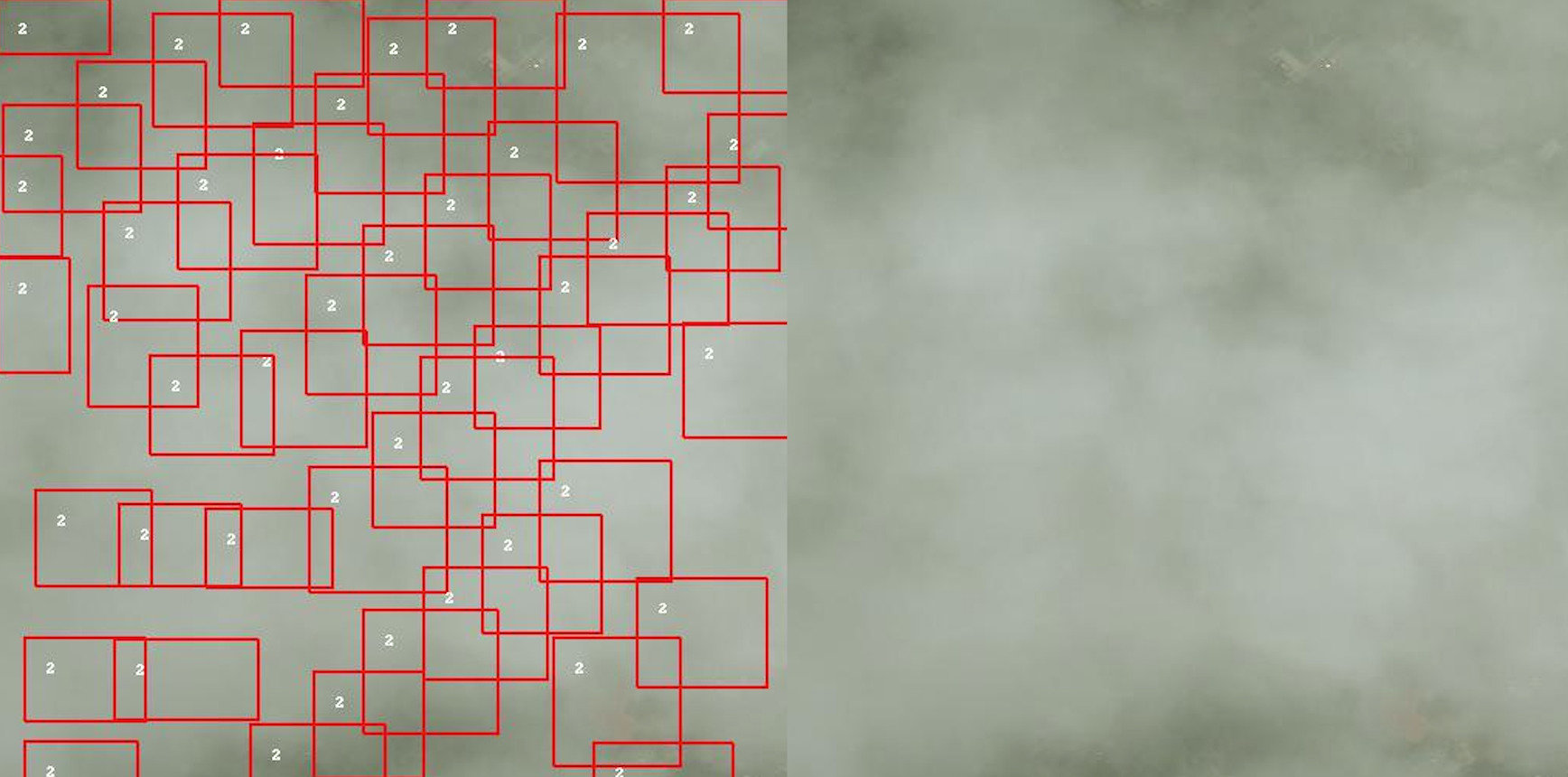}
          \caption{A sample image tile from DigitalGlobe where clouds significantly cover labeled data (red bounding boxes).}
          \label{fig:clouds}
    \end{figure}
     
     As an alternative to the TOMNOD data and the DigitalGlobe data, we also used data created and obtained by the US Government, particularly the Federal Emergency Management Agency (FEMA) and the National Oceanic and Atmospheric Administration (NOAA). Because certain features of the DigitalGlobe satellite imagery could be a source of error for downstream modeling, including missing bands seen in Figure~\ref{fig:bw} and significant cloud cover seen in Figure~\ref{fig:clouds}, we also cleaned and prepared aerial imagery taken by NOAA over several days after Hurricane Harvey. This imagery is below cloud cover and is released at a higher resolution and in RGB. Both datasets posed computational challenges due to their large size; the NOAA data for the affected areas total near 400 gigabytes with a resolution of 9351x9351. To deal with data at such scale, we indexed each raster image as a vector polygon, with which we were able to index damage points to raster images. In the NOAA imagery, many areas overlapped and differed by the date the imagery was acquired. In this case, we use the imagery that is acquired earliest that might evince greater damage upon visual inspection. 
     
     For ground truth data, we also obtained FEMA-estimated flood damages. This dataset was much more comprehensive, linked directly to parcel centroids. FEMA - after a flood event - will use flood maps to estimate flood stages across areas. Depending on the flood stage, FEMA will label different buildings with different degrees of damage, ranging from no damage to major damage. More information on the FEMA methodology can be found here: \url{https://www.fema.gov/media-library/assets/documents/109040}. In many cases, initial estimations are confirmed by on the ground local and federal resources. This dataset was advantageous in that it was more spatially comprehensive across the affected region, and related damages to individual tax parcels. However, the data relate estimated damages which were not necessarily confirmed on the ground. Additionally, damage registered by the dataset may not be visible aerially. However, we assume that damages reported in the dataset are proxies for damages that may be aerially observed. The data provides five classes of damage, including a ``non-damaged" class, explicitly denoting buildings that did not sustain damage, which constitutes an advantage over the TOMNOD dataset. Because the original data was released as the centroids of tax parcels, we acquired county parcel polygon data for all affected counties to which we joined the FEMA dataset. In many cases, the parcel polygons were publicly available online. Other cases required correspondence and official permission, and their distribution is restricted. As such, we do not release the parcel polygons used to create the benchmark datasets that we release. Upon acquiring the parcel polygons, we spatially joined centroids to parcels and in turn spatially joined parcels to building footprints within parcels. In the vast majority of parcels, there was only one structure footprint within the parcel. Very few parcels contained more than one building footprint, in which case the largest footprint was selected. After the footprint was identified, a minimum bounding box was again created, each attributed with level of damage.


\begin{figure}[!ht]
   \centering
   \begin{subfigure}{\linewidth}
   \includegraphics[width=\linewidth]{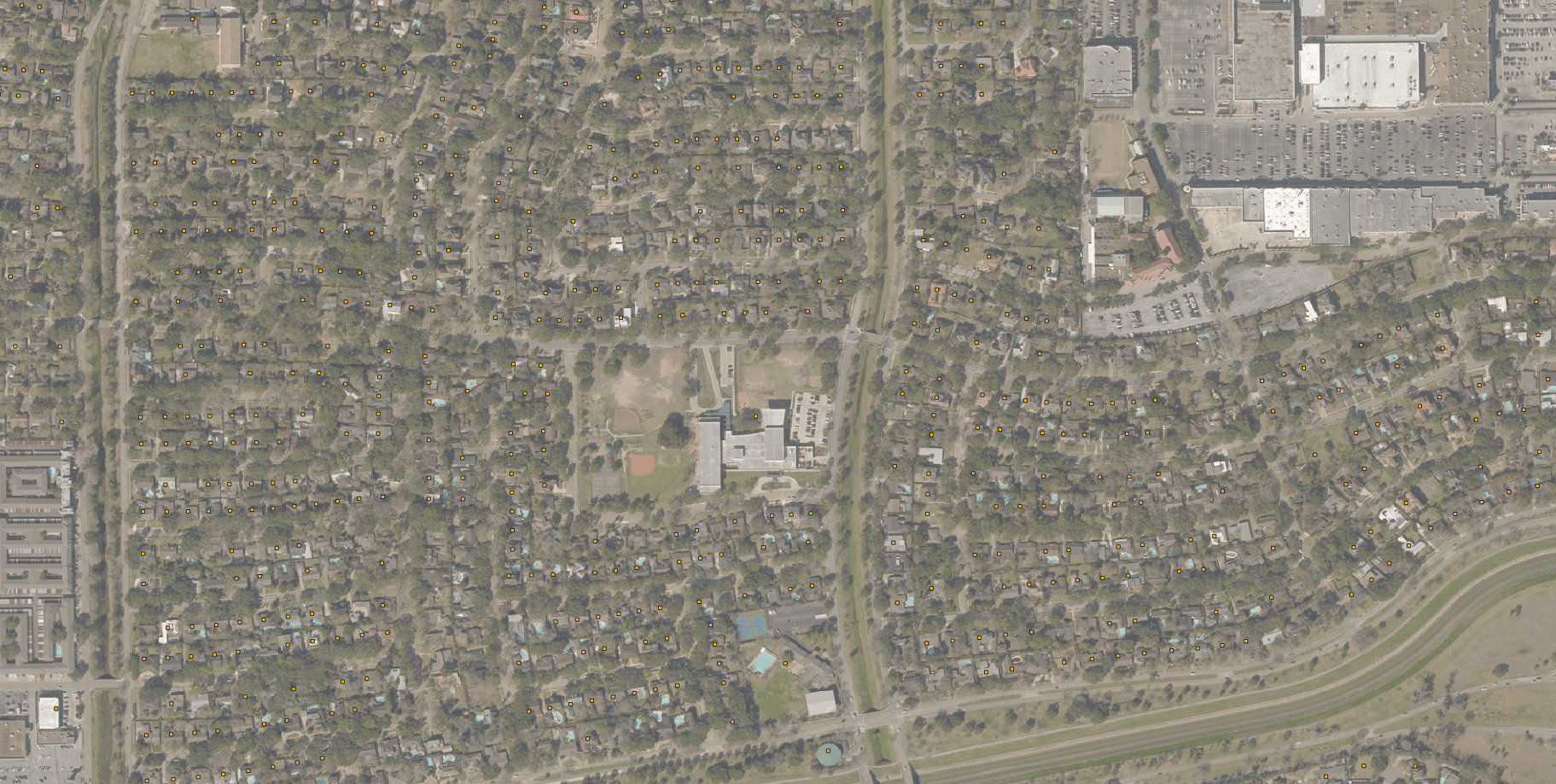}
      \caption{FEMA points}
   \end{subfigure}

      \begin{subfigure}{{\linewidth}}
   \includegraphics[width=\linewidth]{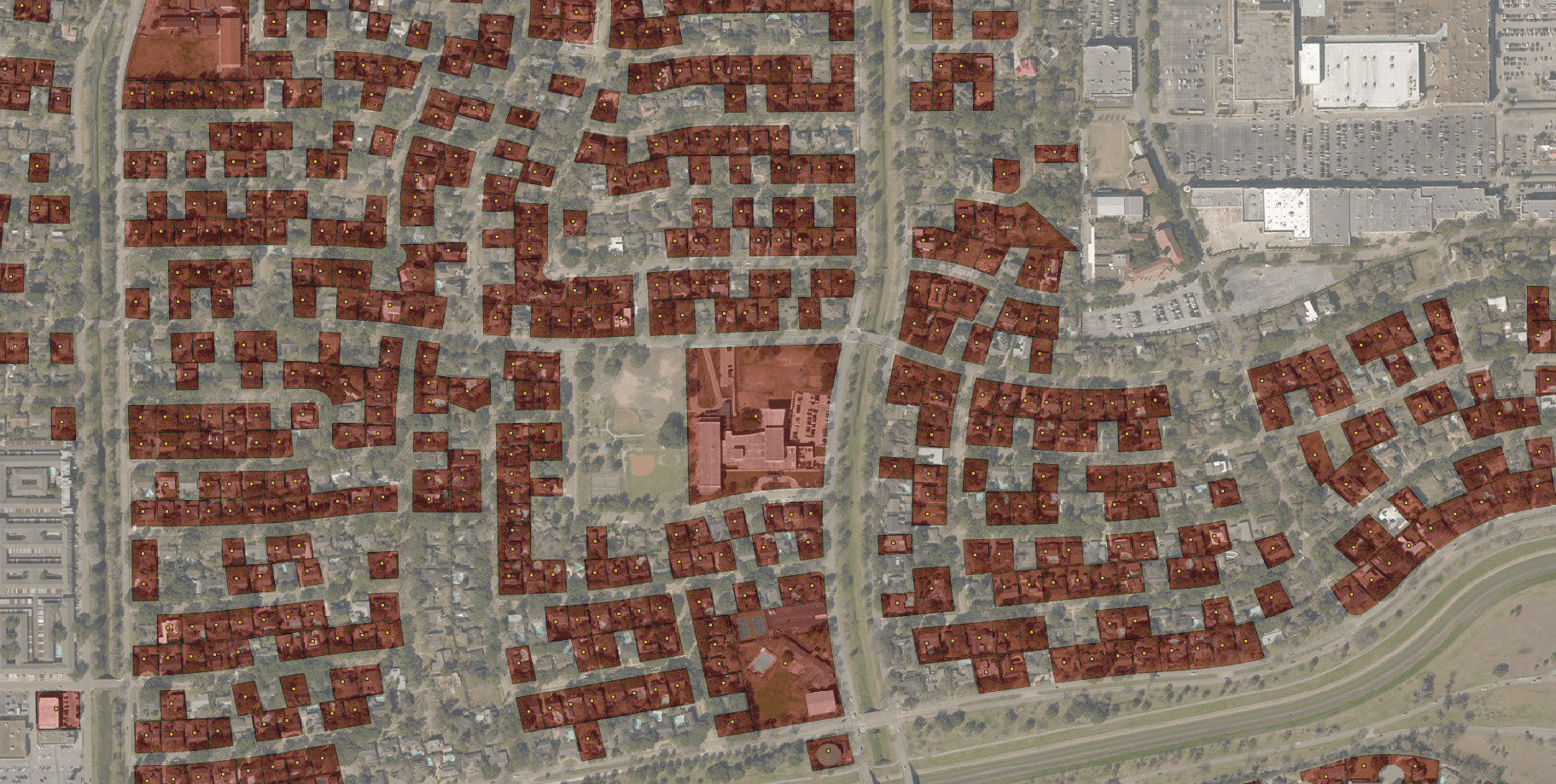}
      \caption{Identified affected parcels}
   \end{subfigure}

      \begin{subfigure}{{\linewidth}}
   \includegraphics[width=\linewidth]{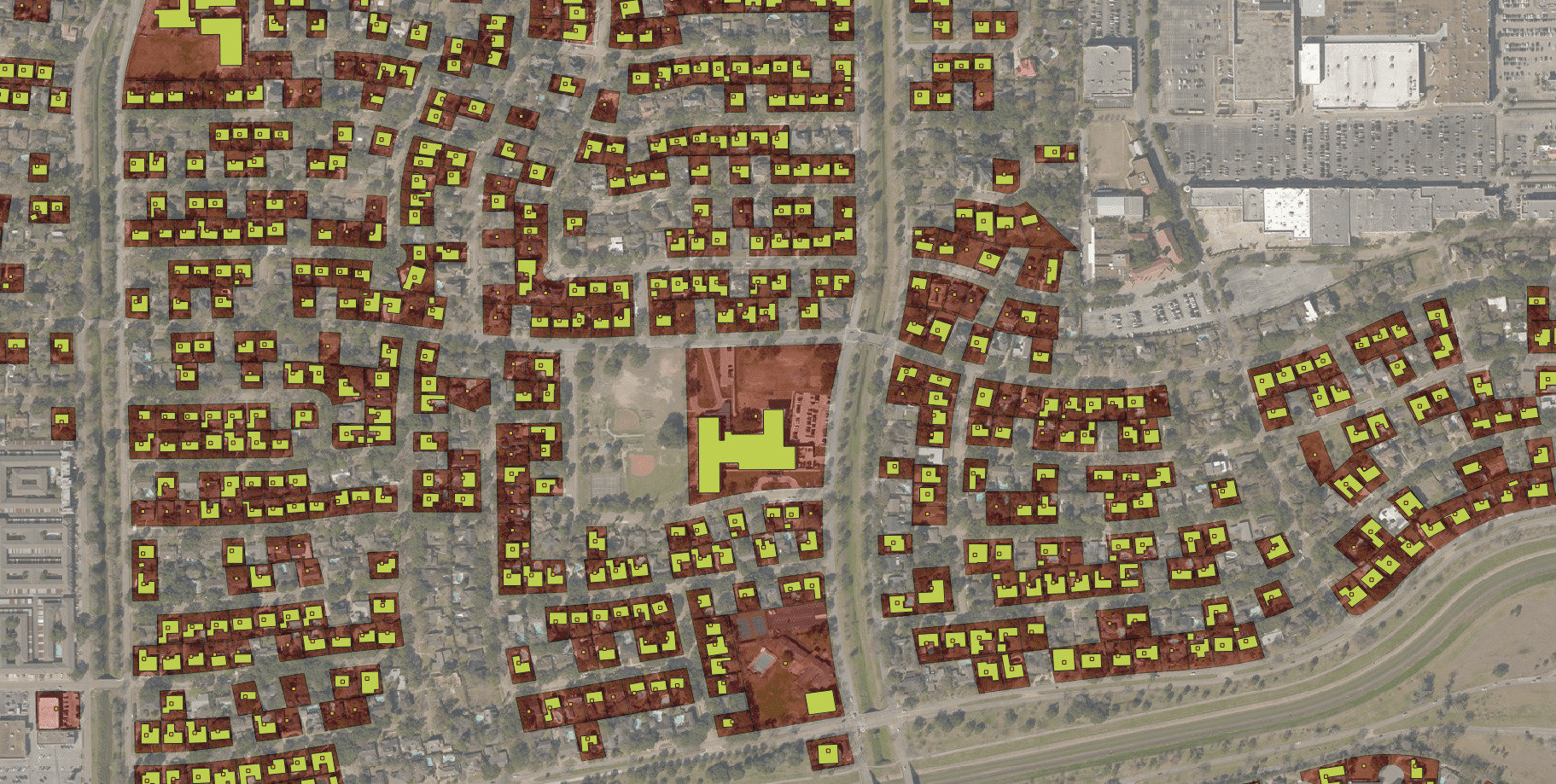}
    \caption{Identified individual structures}
   \end{subfigure}
  
      \begin{subfigure}{{\linewidth}}
   \includegraphics[width=\linewidth]{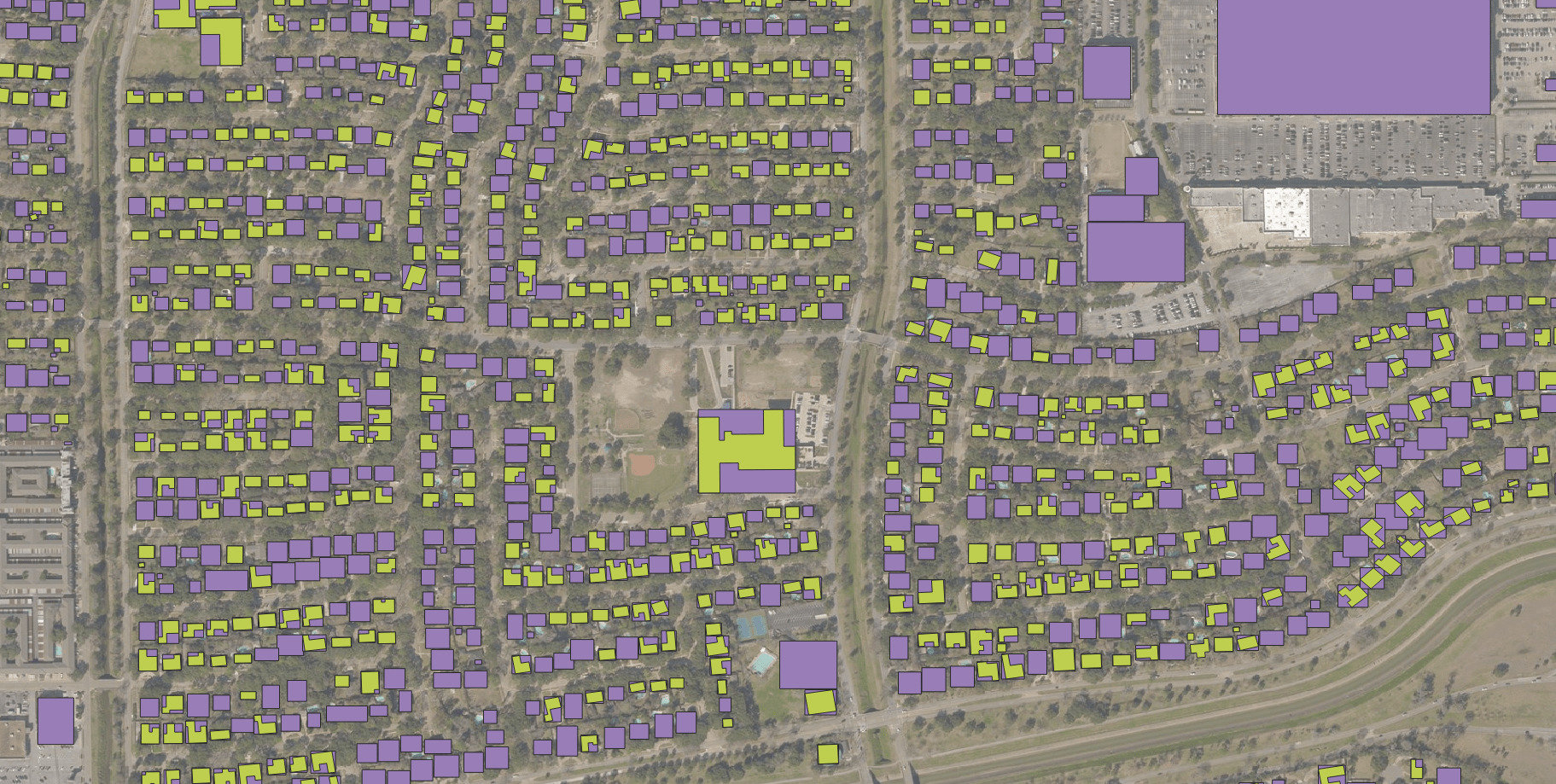}
      \caption{Bounding boxes from identified structures}
   \end{subfigure}
   
  \caption{Process of obtaining bounding boxes from FEMA damage points}

\end{figure}


    \begin{figure}

      \centering
        \includegraphics[width=0.5\textwidth]{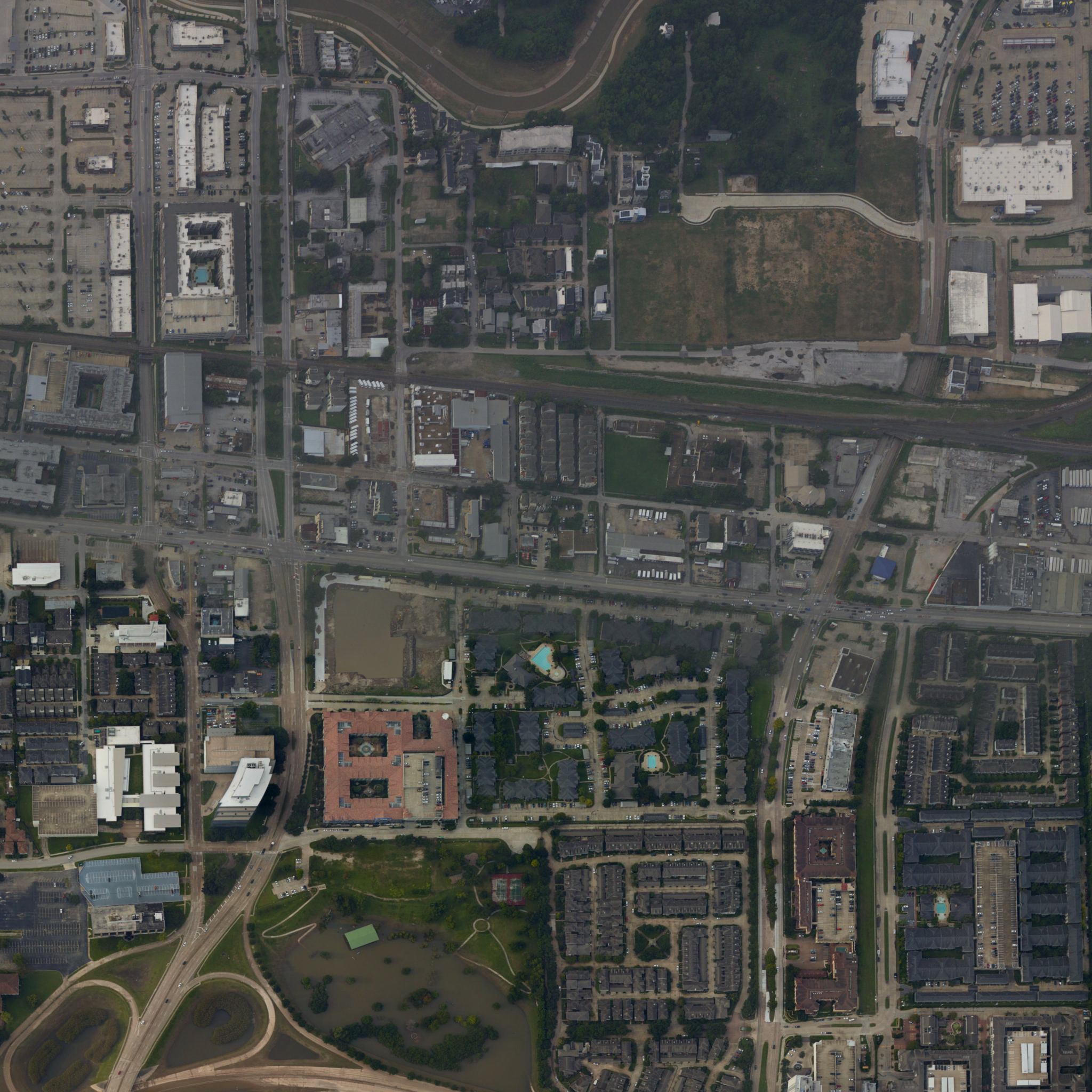}
              \caption{A sample image tile from NOAA.}
    \end{figure}

     To aid in geocomputation times, we loaded vector data into a PostGIS database allowing spatial indexing to decrease computational time and resources. Imagery was also further tiled to create smaller sample sizes as with the more comprehensive FEMA data almost all buildings in the affected areas were labeled.
     
\begin{figure*}[tb]
      \centering
        \includegraphics[width=1\textwidth]{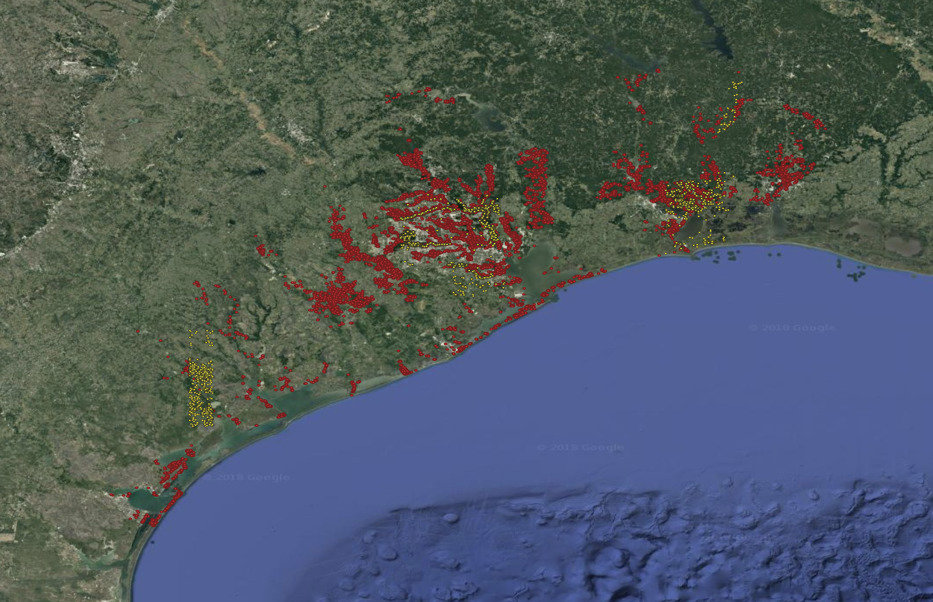}
          \caption{Comparison of labeled FEMA data (red) versus the labeled TOMNOD data (yellow). The number of FEMA points is higher, however, the two dasasets cover different regions and both can be useful for algorithm training.}
          \label{fig:comparison}
    \end{figure*}

     Figure~\ref{fig:comparison} depicts the comparison of labeled FEMA data (red) versus the labeled TOMNOD data (yellow). It can be clearly seen here that the recorded FEMA points significantly outnumber the TOMNOD points. 
     
\section{Conclusion}\label{sec:conclusion} 

In summary, this paper presents a scalable framework to create damaged building datasets that can be used to train object detection models. Our framework consists of largely automated steps with a minimal level of manual intervention (i.e., crowdsourcing of web-based damage annotations and visual inspection of automatically created bounding boxes). To facilitate further research, the benchmark datasets and code created in this work are publicly shared (see Appendix).

Future work includes extensive empirical tests of state-of-the-art object detection methods on the benchmark datasets. 
The shared dataset is in a format to which one could easily apply a wide range of algorithms such as Single Shot MultiBox Detector (SSD)\cite{Wei2016}, Faster R-CNN\cite{Ren2015}, RetinaNet\cite{Lin2017}, etc., without substantial remote sensing knowledge, and without the need of high performance computing resources.
The proposed data development framework can be applied to other hurricane events to expand into a suite of damaged building datasets in a hope to improve the generalizability of automatic damage detection models. 




%

\appendices
\section*{Appendix: Data and Code}

Our benchmark datasets are publicly available through the IEEE DataPort and the following DOI: 
\mbox{\url{https://dx.doi.org/10.21227/1s3n-f891}}. 
The data are stored as ESRI Shapefiles and GeoTIFFs. 


\section*{Acknowledgments}

The authors would like to thank the eScience Institute for the support of this project through the Data Science for Social Good program at the University of Washington. We also thank An Yan who contributed to the project discussion during the program period. 
Quoc Dung Cao's open-source code was helpful for initial steps of data processing. We would like to acknowledge the financial support from the Gordon and Betty Moore and Alfred P. Sloan Foundations, and Microsoft. Y. Choe's work was partially supported by the National Science Foundation (CMMI-1824681). 

The authors gratefully acknowledge DigitalGlobe for data sharing through their Open Data Program (\url{https://www.digitalglobe.com/opendata}), as well as NOAA and FEMA for publicly sharing data on their websites. The authors would also like to thank the following Texas counties for sharing parcel data that was used in preparing our benchmark dataset: Harris County, Orange County, Fort Bend County, Jefferson County, Montgomery County, Wharton County, Galveston County, Nueces County, Aransas County, Victoria County, Matagorda County, Brazoria County, Calhoun County, and Newton County.

\ifCLASSOPTIONcaptionsoff
  \newpage
\fi



\bibliographystyle{IEEEtran}
\bibliography{IEEEabrv,RS_BIB}

\begin{thebibliography}{10}
\providecommand{\url}[1]{#1}
\csname url@samestyle\endcsname
\providecommand{\newblock}{\relax}
\providecommand{\bibinfo}[2]{#2}
\providecommand{\BIBentrySTDinterwordspacing}{\spaceskip=0pt\relax}
\providecommand{\BIBentryALTinterwordstretchfactor}{4}
\providecommand{\BIBentryALTinterwordspacing}{\spaceskip=\fontdimen2\font plus
\BIBentryALTinterwordstretchfactor\fontdimen3\font minus
  \fontdimen4\font\relax}
\providecommand{\BIBforeignlanguage}[2]{{%
\expandafter\ifx\csname l@#1\endcsname\relax
\typeout{** WARNING: IEEEtran.bst: No hyphenation pattern has been}%
\typeout{** loaded for the language `#1'. Using the pattern for}%
\typeout{** the default language instead.}%
\else
\language=\csname l@#1\endcsname
\fi
#2}}
\providecommand{\BIBdecl}{\relax}
\BIBdecl

\bibitem{FEMA2016}
{Federal Emergency Management Agency}, ``Damage assessment operations manual,''
  The U.S. Department of Homeland Security, Tech. Rep., Apr. 2016.

\bibitem{duda2011usgs}
K.~A. Duda and B.~K. Jones, ``{USGS} remote sensing coordination for the 2010
  {Haiti} earthquake,'' \emph{Photogrammetric Engineering \& Remote Sensing},
  vol.~77, no.~9, pp. 899--907, 2011.

\bibitem{ofli2016combining}
F.~Ofli, P.~Meier, M.~Imran, C.~Castillo, D.~Tuia, N.~Rey, J.~Briant,
  P.~Millet, F.~Reinhard, M.~Parkan, and S.~Joost, ``Combining human computing
  and machine learning to make sense of big (aerial) data for disaster
  response,'' \emph{Big Data}, vol.~4, no.~1, pp. 47--59, 2016.

\bibitem{barrington2011crowdsourcing}
L.~Barrington, S.~Ghosh, M.~Greene, S.~Har-Noy, J.~Berger, S.~Gill, A.~Y.-M.
  Lin, and C.~Huyck, ``Crowdsourcing earthquake damage assessment using remote
  sensing imagery,'' \emph{Annals of Geophysics}, vol.~54, no.~6, 2011.

\bibitem{simonite_2018}
\BIBentryALTinterwordspacing
T.~Simonite, ``The {Pentagon} wants your help analyzing satellite images,''
  \emph{WIRED}, Feb 2018. [Online]. Available:
  \url{https://www.wired.com/story/the-pentagon-wants-your-help-analyzing-satellite-images/}
\BIBentrySTDinterwordspacing

\bibitem{mnih_2013}
V.~Mnih, ``Machine learning for aerial image labeling,'' Ph.D. dissertation,
  University of Toronto, 2013.

\bibitem{pohl_2018}
\BIBentryALTinterwordspacing
J.~Pohl, ``Thousands of {FEMA} rescuers spent more time traveling, awaiting
  orders than on rescues,'' \emph{USA TODAY}, Feb 2018. [Online]. Available:
  \url{https://www.usatoday.com/story/news/nation-now/2018/02/18/fema-rescuers-spent-time-traveling-waiting/350163002/}
\BIBentrySTDinterwordspacing

\bibitem{cheng2016survey}
G.~Cheng and J.~Han, ``A survey on object detection in optical remote sensing
  images,'' \emph{ISPRS Journal of Photogrammetry and Remote Sensing}, vol.
  117, pp. 11--28, 2016.

\bibitem{xia2017aid}
G.-S. Xia, J.~Hu, F.~Hu, B.~Shi, X.~Bai, Y.~Zhong, L.~Zhang, and X.~Lu,
  ``{AID}: A benchmark data set for performance evaluation of aerial scene
  classification,'' \emph{IEEE Transactions on Geoscience and Remote Sensing},
  vol.~55, no.~7, pp. 3965--3981, 2017.

\bibitem{cao2018deep}
Q.~D. Cao and Y.~Choe, ``Detecting damaged buildings on post-hurricane
  satellite imagery based on customized convolutional neural networks,''
  \emph{arXiv preprint arXiv:1807.01688}, 2018.

\bibitem{yang2011hierarchical}
Y.~Yang, H.-Y. Ha, F.~Fleites, S.-C. Chen, and S.~Luis, ``Hierarchical disaster
  image classification for situation report enhancement,'' in \emph{Information
  Reuse and Integration (IRI), 2011 IEEE International Conference on}.\hskip
  1em plus 0.5em minus 0.4em\relax IEEE, 2011, pp. 181--186.

\bibitem{hristidis2010survey}
V.~Hristidis, S.-C. Chen, T.~Li, S.~Luis, and Y.~Deng, ``Survey of data
  management and analysis in disaster situations,'' \emph{Journal of Systems
  and Software}, vol.~83, no.~10, pp. 1701--1714, 2010.

\bibitem{corbane2011comparison}
C.~Corbane, D.~Carrion, G.~Lemoine, and M.~Broglia, ``Comparison of damage
  assessment maps derived from very high spatial resolution satellite and
  aerial imagery produced for the {H}aiti 2010 earthquake,'' \emph{Earthquake
  Spectra}, vol.~27, no.~S1, pp. S199--S218, 2011.

\bibitem{lemoine2013intercomparison}
G.~Lemoine, C.~Corbane, C.~Louvrier, and M.~Kauffmann, ``Intercomparison and
  validation of building damage assessments based on post-{H}aiti 2010
  earthquake imagery using multi-source reference data,'' \emph{Natural Hazards
  and Earth System Sciences Discussions}, no.~2, pp. 1445--1486, 2013.

\bibitem{tralli2005satellite}
D.~M. Tralli, R.~G. Blom, V.~Zlotnicki, A.~Donnellan, and D.~L. Evans,
  ``Satellite remote sensing of earthquake, volcano, flood, landslide and
  coastal inundation hazards,'' \emph{ISPRS Journal of Photogrammetry and
  Remote Sensing}, vol.~59, no.~4, pp. 185--198, 2005.

\bibitem{digitalglobe2018}
\BIBentryALTinterwordspacing
DigitalGlobe. (2018, Jul.) Open data program. [Online]. Available:
  \url{https://www.digitalglobe.com/opendata/hurricane-harvey/}
\BIBentrySTDinterwordspacing

\bibitem{planet2018}
\BIBentryALTinterwordspacing
A.~Zolli. (2018, Jul.) Planet announces new, flexible emergency and disaster
  management solution. [Online]. Available:
  \url{https://www.planet.com/pulse/planet-emergency-and-disaster-management-solution/}
\BIBentrySTDinterwordspacing

\bibitem{bielski2011post}
C.~Bielski, S.~Gentilini, and M.~Pappalardo, ``Post-disaster image processing
  for damage analysis using {G}{E}{N}{E}{S}{I}-{D}{R}, {W}{P}{S} and grid
  computing,'' \emph{Remote Sensing}, vol.~3, no.~6, pp. 1234--1250, 2011.

\bibitem{earthnow2018}
\BIBentryALTinterwordspacing
EarthNow. (2018, Jul.) {E}arth{N}ow. [Online]. Available:
  \url{https://earthnow.com/}
\BIBentrySTDinterwordspacing

\bibitem{bialas2016object}
J.~Bialas, T.~Oommen, U.~Rebbapragada, and E.~Levin, ``Object-based
  classification of earthquake damage from high-resolution optical imagery
  using machine learning,'' \emph{Journal of Applied Remote Sensing}, vol.~10,
  no.~3, p. 036025, 2016.

\bibitem{dfo2018}
\BIBentryALTinterwordspacing
{Dartmouth Flood Observatory}. (2018, Jul.) The {D}artmouth {F}lood
  {O}bservatory. [Online]. Available:
  \url{http://floodobservatory.colorado.edu/}
\BIBentrySTDinterwordspacing

\bibitem{science2018}
J.~Plautz, ``In {C}olorado, a global flood observatory keeps a close watch on
  {H}arvey's torrents,'' \emph{Science}, 2017.

\bibitem{medium2017}
\BIBentryALTinterwordspacing
J.~Kwok. (2017) Deep learning for disaster recovery: Automatic detection of
  flooded roads. [Online]. Available:
  \url{https://blog.insightdatascience.com/deep-learning-for-disaster-recovery-45c8cd174d7a}
\BIBentrySTDinterwordspacing

\bibitem{gbdx2018}
\BIBentryALTinterwordspacing
{GBDX Stories}. (2017, Jul.) Unsupervised flood mapping. [Online]. Available:
  \url{http://gbdxstories.digitalglobe.com/flood-water/}
\BIBentrySTDinterwordspacing

\bibitem{planet2}
\BIBentryALTinterwordspacing
Planet. (2018, Jul.) Anatomy of a catastrophe: Using imagery to assess
  {H}arvey's impact on {H}ouston. [Online]. Available:
  \url{https://www.planet.com/insights/anatomy-of-a-catastrophe/}
\BIBentrySTDinterwordspacing

\bibitem{limmer2016infrared}
M.~Limmer and H.~P. Lensch, ``Infrared colorization using deep convolutional
  neural networks,'' in \emph{Machine Learning and Applications (ICMLA), 2016
  15th IEEE International Conference on}.\hskip 1em plus 0.5em minus
  0.4em\relax IEEE, 2016, pp. 61--68.

\bibitem{aria2018}
\BIBentryALTinterwordspacing
{Advanced Rapid Imaging and Analysis}. (2018, Jul.) About {ARIA}. [Online].
  Available: \url{https://aria.jpl.nasa.gov/about}
\BIBentrySTDinterwordspacing

\bibitem{dong2013comprehensive}
L.~Dong and J.~Shan, ``A comprehensive review of earthquake-induced building
  damage detection with remote sensing techniques,'' \emph{ISPRS Journal of
  Photogrammetry and Remote Sensing}, vol.~84, pp. 85--99, 2013.

\bibitem{rathje2005damage}
E.~M. Rathje, M.~Crawford, K.~Woo, and A.~Neuenschwander, ``Damage patterns
  from satellite images of the 2003 {B}am, {I}ran, earthquake,''
  \emph{Earthquake Spectra}, vol.~21, no.~S1, pp. 295--307, 2005.

\bibitem{bignami2011objects}
C.~Bignami, M.~Chini, S.~Stramondo, W.~J. Emery, and N.~Pierdicca, ``Objects
  textural features sensitivity for earthquake damage mapping,'' in \emph{Urban
  Remote Sensing Event (JURSE), 2011 Joint}.\hskip 1em plus 0.5em minus
  0.4em\relax IEEE, 2011, pp. 333--336.

\bibitem{li2011improved}
X.~Li, W.~Yang, T.~Ao, H.~Li, and W.~Chen, ``An improved approach of
  information extraction for earthquake-damaged buildings using high-resolution
  imagery,'' \emph{Journal of Earthquake and Tsunami}, vol.~5, no.~04, pp.
  389--399, 2011.

\bibitem{brunner2010earthquake}
D.~Brunner, G.~Lemoine, and L.~Bruzzone, ``Earthquake damage assessment of
  buildings using {V}{H}{R} optical and {S}{A}{R} imagery,'' \emph{IEEE
  Transactions on Geoscience and Remote Sensing}, vol.~48, no.~5, pp.
  2403--2420, 2010.

\bibitem{lam2018xview}
D.~Lam, R.~Kuzma, K.~McGee, S.~Dooley, M.~Laielli, M.~Klaric, Y.~Bulatov, and
  B.~McCord, ``x{V}iew: Objects in context in overhead imagery,'' \emph{arXiv
  preprint arXiv:1802.07856}, 2018.

\bibitem{Erinjippurath2018}
\BIBentryALTinterwordspacing
G.~Erinjippurath. (2018, Nov.) Accelerating disaster response at the
  intersection of space and ml. [Online]. Available:
  \url{https://medium.com/planet-stories/accelerating-disaster-response-at-the-intersection-of-space-and-ml-ff2fe8ebfa1d}
\BIBentrySTDinterwordspacing

\bibitem{TOMNOD}
{TOMNOD}, ``Hurricane harvey annotations,''
  \url{https://www.digitalglobe.com/opendata/hurricane-harvey/vector-data},
  2018.

\bibitem{FEMAdata}
\BIBentryALTinterwordspacing
F.~E.~M. Agency. (2018, Jul.) Fema modeled building damage assessments harvey.
  [Online]. Available:
  \url{https://respond-harvey-geoplatform.opendata.arcgis.com/datasets/1ac0b7c856a047e6bba2c66b32982f00_0}
\BIBentrySTDinterwordspacing

\bibitem{noaa2018}
\BIBentryALTinterwordspacing
N.~Oceanic and A.~Administration. (2017, Sep.) Hurricane harvey imagery.
  [Online]. Available:
  \url{https://storms.ngs.noaa.gov/storms/harvey/index.html#7/28.400/-96.690}
\BIBentrySTDinterwordspacing

\bibitem{MicrosoftBF}
Microsoft, ``{U.S.} building footprints,''
  \url{https://github.com/Microsoft/USBuildingFootprints}, 2018.

\bibitem{OakRidgeBF}
\BIBentryALTinterwordspacing
O.~R.~N. Laboratory. (2017, Nov.) Post harvey building footprints. [Online].
  Available:
  \url{https://www.ornl.gov/news/datasets-supporting-hurricane-damage-assessments}
\BIBentrySTDinterwordspacing

\bibitem{Wei2016}
W.~Liu, D.~Anguelov, D.~Erhan, C.~Szegedy, S.~Reed, C.-Y. Fu, and A.~C. Berg,
  ``{S}{S}{D}: Single shot multibox detector,'' in \emph{Computer Vision --
  ECCV 2016}, B.~Leibe, J.~Matas, N.~Sebe, and M.~Welling, Eds.\hskip 1em plus
  0.5em minus 0.4em\relax Cham: Springer International Publishing, 2016, pp.
  21--37.

\bibitem{Ren2015}
\BIBentryALTinterwordspacing
S.~Ren, K.~He, R.~Girshick, and J.~Sun, ``Faster {R}-{CNN}: Towards real-time
  object detection with region proposal networks,'' in \emph{Proceedings of the
  28th International Conference on Neural Information Processing Systems -
  Volume 1}, ser. NIPS'15.\hskip 1em plus 0.5em minus 0.4em\relax Cambridge,
  MA, USA: MIT Press, 2015, pp. 91--99. [Online]. Available:
  \url{http://dl.acm.org/citation.cfm?id=2969239.2969250}
\BIBentrySTDinterwordspacing

\bibitem{Lin2017}
T.~Lin, P.~Goyal, R.~Girshick, K.~He, and P.~Dollár, ``Focal loss for dense
  object detection,'' in \emph{2017 IEEE International Conference on Computer
  Vision (ICCV)}, Oct 2017, pp. 2999--3007.

\end{thebibliography}
\end{document}